\documentclass[sigconf]{acmart}
\pdfoutput=1
\acmYear{}
\acmPrice{}
\acmISBN{}
\acmDOI{}
\setcopyright{none}
\acmConference[AdvML'19: Workshop on Adversarial Learning Methods for Machine Learning and Data Mining at KDD]{AdvML'19: Workshop on Adversarial Learning Methods for Machine Learning and Data Mining at KDD}{August 5th, 2019}{Anchorage, Alaska, USA}

\usepackage{subcaption}
\usepackage{tikz}
\usetikzlibrary{arrows,calc}
\usepackage{multirow}

\begin{document}

\title{Multi-Purposing Domain Adaptation Discriminators for Pseudo Labeling
Confidence}

\author{Garrett Wilson}
\email{garrett.wilson@wsu.edu}
\orcid{0000-0002-6760-754X}
\affiliation{%
  \institution{Washington State University}
  \streetaddress{School of Electrical Engineering and Computer Science}
  \city{Pullman}
  \state{WA}
  \postcode{99164}
}

\author{Diane J. Cook}
\email{djcook@wsu.edu}
\orcid{0000-0002-4441-7508}
\affiliation{%
  \institution{Washington State University}
  \streetaddress{School of Electrical Engineering and Computer Science}
  \city{Pullman}
  \state{WA}
  \postcode{99164}
}

\begin{abstract}
Often domain adaptation is performed using a discriminator (domain classifier) to learn domain-invariant feature representations so that a classifier trained on labeled source data will generalize well to unlabeled target data. A line of research stemming from semi-supervised learning uses pseudo labeling to directly generate ``pseudo labels'' for the unlabeled target data and trains a classifier on the now-labeled target data, where the samples are selected or weighted based on some measure of confidence. In this paper, we propose multi-purposing the discriminator to not only aid in producing domain-invariant representations but also to provide pseudo labeling confidence.
\end{abstract}



\maketitle

\section{Introduction}
Unsupervised domain adaptation is a problem consisting of two domains: a source domain and a target domain. Labeled source data and unlabeled target data are available for use during training, and the goal is to learn a model that performs well on data from the target domain \cite{pan2010tkde,goodfellow2016deep,ganin2016jmlr}. As a result, this can be used to reduce the need for costly labeled data in the target domain.

A common approach for domain adaptation is to learn a domain-invariant feature representation, which in deep learning methods is typically a feature extractor neural network. Intuitively, if a classifier trained on these domain-invariant features of the labeled source data performs well, then the classifier may generalize to the unlabeled target data since the feature distributions for both domains will be highly similar. (Though, performance on the target data depends on how similar the domains are, and this method may actually increase the error if the domains are too different \cite{ben2010ml,zhao2019learning}.) Numerous methods proposed for achieving this goal have yielded promising results, and many of these methods use adversarial training \cite{wilson2019survey}.

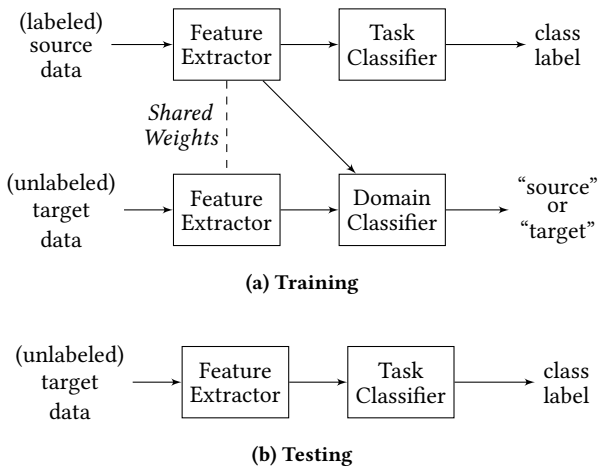
\begin{figure}
  \tikzstyle{int}=[draw, minimum size=3em]
  \tikzstyle{init} = [pin edge={to-,thin,black}]

  \begin{subfigure}{1.0\linewidth}
      \centering
      \begin{tikzpicture}[node distance=2.2cm,auto,>=latex',scale=1.0,every node/.style={scale=1.0}]
          \node (a) {\shortstack{(labeled) \\ source \\ data}};
          \node (b) [below of=a] {\shortstack{(unlabeled) \\ target \\ data}};
          \node [int] (c) [right of=a] {\shortstack{Feature \\ Extractor}};
          \node [int] (d) [right of=b] {\shortstack{Feature \\ Extractor}};
          \node [int] (e) [right of=c] {\shortstack{Task \\ Classifier}};
          \node [int] (f) [right of=d] {\shortstack{Domain \\ Classifier}};
          \node (g) [right of=e] {\shortstack{class \\ label }};
          \node (h) [right of=f] {\shortstack{``source'' \\ or \\ ``target''}};
          \path[->] (a) edge node {} (c);
          \path[->] (b) edge node {} (d);
          \path[->] (c) edge node {} (e);
          \path[->] (c) edge node {} (f);
          \path[->] (d) edge node {} (f);
          \path[->] (e) edge node {} (g);
          \path[->] (f) edge node {} (h);
          \draw[dashed,-] (c) -- (d) node [midway, left] (textNode1) {\shortstack{\textit{Shared} \\ \textit{Weights}}};
      \end{tikzpicture}
      \caption{Training}\label{fig:dann_a}
      \vspace*{5mm}
  \end{subfigure}
  \begin{subfigure}{1.0\linewidth}
      \centering
      \begin{tikzpicture}[node distance=2.2cm,auto,>=latex',scale=1.0,every node/.style={scale=1.0}]
          \node (a) {\shortstack{(unlabeled) \\ target \\ data}};
          \node [int] (b) [right of=a] {\shortstack{Feature \\ Extractor}};
          \node [int] (c) [right of=b] {\shortstack{Task \\ Classifier}};
          \node (d) [right of=c] {\shortstack{class \\ label }};
          \path[->] (a) edge node {} (b);
          \path[->] (b) edge node {} (c);
          \path[->] (c) edge node {} (d);
      \end{tikzpicture}
      \caption{Testing}\label{fig:dann_b}
  \end{subfigure}
  \caption{Network setup for DANN that learns a domain-invariant feature representation -- the basis for our proposed method.}
  \Description{Network setup for DANN consists of three networks: a feature extractor followed by both a task classifier and a domain classifier. At test-time, the feature extractor is followed by the task classifier. This is the basis for our proposed method.}
  \label{fig:dann}
\end{figure}

One such adversarial domain-invariant feature learning method is the domain-adversarial neural network (DANN) \cite{ganin2015icml,ganin2016jmlr}, which is a typical baseline for other variants. This method consists of a feature extractor network followed by two additional networks: a task classifier and a domain classifier (Figure~\ref{fig:dann}). The network is updated by two competing objectives: (1) the feature extractor followed by the task classifier learns to correctly classify the labeled source data while the domain classifier learns to correctly predict whether the features originated from source or target data, and (2) the feature extractor learns to make the domain classifier predict the domain incorrectly. To this end, they propose a gradient reversal layer between the feature extractor and domain classifier so that during backpropagation, the gradient is negated when updating the feature extractor weights. More recently, Shu et al. \cite{shu2018vada} found replacing the gradient reversal layer with adversarial alternating updates from generative adversarial networks (GANs) \cite{goodfellow2014nips} to perform better.

Pseudo labeling is a technique from semi-supervised learning that is also sometimes included in domain-invariant feature learning methods for domain adaptation \cite{saito2017icml,zou2018eccv,das2018graphmatching,sener2016nips}. In pseudo labeling, a source classifier trained on the labeled source data is first used to label the unlabeled target data, generating ``pseudo'' labels that may not all be correct. Next, a target classifier can be trained in a supervised manner on the now-labeled target data. Often there is a selection criterion to utilize only pseudo-labeled data that are more-likely correct (i.e., the model is more confident on those samples). Typically the selection is based on whether the softmax output prediction entropy is low enough \cite{oliver2018nips,zou2018eccv,das2018graphmatching}. The softmax output can be viewed as a probability distribution over the possible labels, so a uniform distribution over these predictions indicates the model has no idea what label to predict whereas a very high prediction probability for one class (low entropy) indicates the model has high confidence in a prediction. Other measures of confidence include ensemble agreement (combined with softmax confidence) \cite{saito2017icml} or $k$-nearest neighbor agreement \cite{sener2016nips}.

In this paper, we propose another selection criterion for pseudo labeling: the discriminator's confidence. Methods such as DANN already have a discriminator, allowing it to be easily multi-purposed to not only aid in producing domain-invariant representations but also to provide pseudo labeling confidence. The domain discriminator learns to classify feature representations as either source domain or target domain, but in unsupervised domain adaptation this can also be interpreted as known label vs. unknown label, or rather, accurate vs. possibly inaccurate, assuming the task classifier performs well on the source data. Thus, we could view samples as ``confident'' if the discriminator incorrectly classifies the target samples' feature representations as originating from the source domain. Intuitively, this process may select samples that are pseudo labeled correctly since the feature representation was close to that of data with known labels. However, our proposed approach assumes (1) the task classifier does perform well on the labeled source data and (2) there exists sufficient similarity between domains. The first assumption is easy to verify during training. The second assumption is harder to quantify, but empirically we obtain high target domain performance.

To explain our proposed method, we first discuss the relationship with existing methods. Second, we describe our method in detail. Finally, we perform experiments on a variety of image datasets commonly used for domain adaptation.

\section{Related Work}
Numerous domain-invariant feature learning methods have been proposed. Some do this by minimizing a divergence such as maximum mean discrepancy \cite{rozantsev2018ieee,long2015icml,long2017jmmd}, second-order statistics \cite{sun2016,zhang2018mca,wang2017iccv,morerio2017correlation,morerio2018minimalentropy}, or contrastive domain discrepancy \cite{kang2019contrastive}. Others use optimal transport \cite{damodaran2018deepjdot,courty2017nips}, graph matching \cite{das2018graphmatching}, or reconstruction \cite{ghifary2016}. Still others learn domain-invariant features adversarially with a domain classifier \cite{ajakan2014domain,ganin2015icml,ganin2016jmlr,purushotham2017variational,tzeng2015iccv,tzeng2017cvpr,long2018nips,shen2018wasserstein} or a GAN \cite{sankaranarayanan2018cvpr,sankaranarayanan2018cvprsemantic}. The method in this paper is based on DANN, an adversarial approach.

Several domain-invariant adaptation methods also incorporate pseudo labeling to further improve performance. Some select confident samples that have low entropy \cite{zou2018eccv,das2018graphmatching}. Others use an ensemble of networks that make independent predictions and select confident samples based on a combination of the ensemble agreement and verifying that at least one of the ensemble predictions has low entropy \cite{saito2017icml}. One method classifies with $k$-nearest neighbors and thus bases its confidence on agreement of the $k$ predictions \cite{sener2016nips}. In this paper, we propose using the DANN discriminator to provide a measure of confidence for pseudo labeling.

Pseudo labeling can be viewed as conditional entropy regularization \cite{lee2013pseudo,grandvalet2004nips}, which while proposed for semi-supervised learning has also been applied in domain adaptation methods \cite{shu2018vada,kumar2018nips}. Entropy regularization and pseudo labeling are based upon the cluster assumption: data are clustered by class/label and separated by low-density regions. If this is true, then decision boundaries should lie in these low-density regions \cite{chapelle2005semi,lee2013pseudo}. Entropy regularization is one way to move decision boundaries away from regions with higher density. However, this assumes that the decisions do not drastically change when approaching data points, i.e., that the model is locally Lipschitz \cite{shu2018vada}. This can be enforced with virtual adversarial training \cite{miyato2018virtual}, which thus is typically also used when applying entropy regularization to domain adaptation \cite{shu2018vada,kumar2018nips}. Alternative methods have also been proposed with the same effect of moving decision boundaries into lower-density regions based on GANs \cite{wei2018generative}, adversarial dropout \cite{saito2018adversarial}, and self-ensembling \cite{tarvainen2017nips,laine2017iclr,french2018iclr}. We base our method on pseudo labeling rather than entropy regularization or the alternative methods.

Pseudo labeling is related to self-training and expectation maximization. In self-training, a classifier is trained on labeled data, predicts labels of unlabeled data, and is re-trained on the previously-unlabeled data. This process is then repeated \cite{zhu2005semi}. Self-training can be shown to be equivalent to a particular classification expectation maximization algorithm \cite{grandvalet2004nips,amini2002semi}. Pseudo labeling is almost the same, except that it is trained simultaneously on labeled and unlabeled data \cite{lee2013pseudo}. In domain adaptation, we have two separate domains, so we may instead wish to use the pseudo-labeled target data to train a separate target classifier \cite{saito2017icml}. This could be done in either one or two steps (similar to pseudo labeling or self-training, respectively).

Pseudo labeling is also related to co-training. Co-training is similar to self-training but utilizes two classifiers for two separate views of the data. Pseudo-labeled samples are selected in which exactly one of the classifiers is confident, which are then added to the labeled training set for training in subsequent iterations \cite{chen2011nips}. When only one view is available as is common in domain adaptation problems, Chen et al. \cite{chen2011nips} propose feature splits to artificially create two views.

Finally, the proposed method of using a discriminator or domain classifier to select which samples to use for adaptation is related to selection adaptation, a type of instance weighting \cite{beijbom2012domain,daume2012course}. In selection adaptation, a domain classifier learns to predict which domain the samples are from. The labeled source data weighted by a function of these domain predictions are used to train a target classifier \cite{daume2012course}. While related, this differs from our proposed method in several ways. First, we do not weight the source data but rather pseudo label and weight the target data. Second, our discriminator operates at a feature level rather than a sample level. Third, our discriminator is trained jointly rather than in stages (more common in deep methods).

\section{Method}
We compare several alternative approaches for domain adaptation. In the first approach, no adaptation is performed. In the second method, we use DANN to learn a domain-invariant feature representation. The third approach employs pseudo labeling and weights instances either by the task classifier's softmax confidence or by a discriminator's confidence (our proposed method).

\subsection{No adaptation}
We train a feature extractor followed by a task classifier on the labeled source data only. Then we evaluate this model on the target data to see how well it generalizes without performing any domain adaptation. We expect this method to perform poorly when large differences exist between domains.

\subsection{DANN}
We train a feature extractor, softmax task classifier, and binary domain classifier as shown in Figure~\ref{fig:dann}. The training consists of three weight updates at each iteration: (1) the feature extractor and task classifier together are trained to correctly classify labeled source data (e.g., with categorical cross entropy loss), (2) the domain classifier is trained to correctly label from which domain's data the feature representation originated, and (3) the feature extractor is trained to fool the domain classifier. Through this process, the feature extractor learns to produce domain-invariant representations.

Rather than using a gradient reversal layer \cite{ganin2016jmlr} for steps (2) and (3), we choose to perform a GAN-like update as used by Shu et al. \cite{shu2018vada}. For a discriminator $D$, a feature extractor $F$, source domain data $\mathcal{D}_s$, and target domain data $\mathcal{D}_t$, these two updates can be performed by minimizing:

\begin{equation}\label{eq:step1}
\min_D - \mathbb{E}_{x \sim \mathcal{D}_s} \left[ \log D(F(x)) \right]
    - \mathbb{E}_{x \sim \mathcal{D}_t} \left[ \log(1 - D(F(x))) \right]
\end{equation}
\begin{equation}\label{eq:step2}
\min_F - \mathbb{E}_{x \sim \mathcal{D}_t} \left[ \log D(F(x)) \right]
    - \mathbb{E}_{x \sim \mathcal{D}_s} \left[ \log(1 - D(F(x))) \right]
\end{equation}

Step (2) becomes Equation~\ref{eq:step1}, updating the discriminator to correctly classify the feature representations of source and target data as ``source'' and ``target''. Step (3) becomes Equation~\ref{eq:step2}, updating the feature extractor to fool the discriminator by classifying source data as ``target'' and target data as ``source''. The losses for these two updates can be computed with binary cross entropy.

\subsection{Pseudo labeling}
\begin{figure}
  \tikzstyle{int}=[draw, minimum size=3em]
  \tikzstyle{init} = [pin edge={to-,thin,black}]

  \begin{subfigure}{1.0\linewidth}
      \centering
      \begin{tikzpicture}[node distance=2.2cm,auto,>=latex',scale=1.0,every node/.style={scale=1.0}]
          \node (a) {\shortstack{(unlabeled) \\ target \\ data}};
          \node [int] (b) [right of=a] {\shortstack{Feature \\ Extractor}};
          \node [int] (c) [right of=b] {\shortstack{Task \\ Classifier}};
          \node [int] (d) [below of=c] {\shortstack{Domain \\ Classifier}};
          \node (e) [right of=c] {\shortstack{pseudo \\ label }};
          \node (f) [right of=d] {\shortstack{``source'' \\ or \\ ``target''}};
          \path[->] (a) edge node {} (b);
          \path[->] (b) edge node {} (c);
          \path[->] (b) edge node {} (d);
          \path[->] (c) edge node {} (e);
          \path[->] (d) edge node {} (f);
      \end{tikzpicture}
      \caption{Training: pseudo label step (no weight update)}\label{fig:pseudo_a}
      \vspace*{5mm}
  \end{subfigure}
  \begin{subfigure}{1.0\linewidth}
    \centering
    \begin{tikzpicture}[node distance=2.2cm,auto,>=latex',scale=1.0,every node/.style={scale=1.0}]
        \node (a) {\shortstack{(unlabeled) \\ target \\ data}};
        \node [int] (b) [right of=a] {\shortstack{Feature \\ Extractor}};
        \node [int] (c) [right of=b] {\shortstack{Target \\ Classifier}};
        \node (d) [right of=c] {\shortstack{(pseudo- \\ labeled) \\ class label }};
        \path[->] (a) edge node {} (b);
        \path[->] (b) edge node {} (c);
        \path[->] (c) edge node {} (d);
    \end{tikzpicture}
    \caption{Training: update target classifier step}\label{fig:pseudo_b}
    \vspace*{5mm}
\end{subfigure}
  \begin{subfigure}{1.0\linewidth}
      \centering
      \begin{tikzpicture}[node distance=2.2cm,auto,>=latex',scale=1.0,every node/.style={scale=1.0}]
          \node (a) {\shortstack{(unlabeled) \\ target \\ data}};
          \node [int] (b) [right of=a] {\shortstack{Feature \\ Extractor}};
          \node [int] (c) [right of=b] {\shortstack{Target \\ Classifier}};
          \node (d) [right of=c] {\shortstack{class \\ label }};
          \path[->] (a) edge node {} (b);
          \path[->] (b) edge node {} (c);
          \path[->] (c) edge node {} (d);
      \end{tikzpicture}
      \caption{Testing}\label{fig:pseudo_c}
  \end{subfigure}
  \caption{After each DANN training step (Figure~\ref{fig:dann_a}), target data is (a) pseudo labeled, and then (b) a target classifier is trained on those pseudo labels but weighted by the discriminator's predictions from (a), representing the probability the feature representation was generated from \textit{source} data (i.e., the discriminator was fooled). All feature extractors share weights. At test time, the feature extractor and target classifier are used for making predictions. Note that DANN uses a \textit{task classifier} whereas the above \textit{target classifier} is a separate classifier and is only trained on pseudo-labeled target data.}
  \Description{}
  \label{fig:pseudo}
\end{figure}
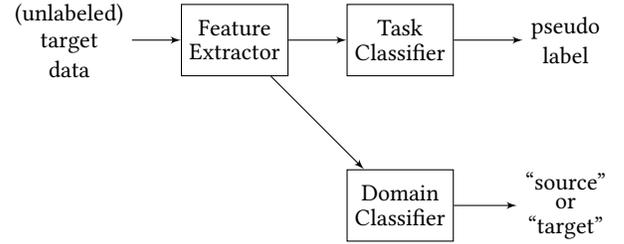
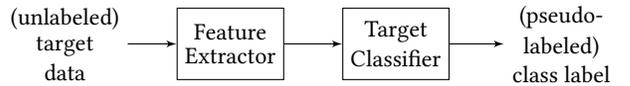
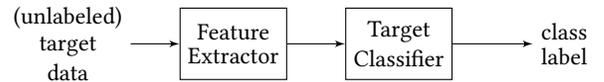

Pseudo labeling can be added to DANN using the following steps. First, perform updates to the feature extractor, task classifier, and domain classifier on a batch of source and target data as in DANN (Figure~\ref{fig:dann_a}). Second, pseudo label a batch of target data using the \textit{task classifier} and record the domain classifier predictions without updating the model (Figure~\ref{fig:pseudo_a}). Third, train a \textit{target classifier} on this pseudo-labeled target data but weighted by the probability that the feature representations were generated from source data (Figure~\ref{fig:pseudo_b}). If the domain classifier is a binary classifier where 0 is ``source'' and 1 is ``target'', this probability can be calculated as $1 - D(F(x))$. This contrasts with weighting by the task classifier's max softmax output probability as a measure of confidence. Both methods of weighting are evaluated in the experiments.

\subsection{Instance weighting}
Alternatively, we can replace pseudo labeling with instance weighting. In this case we train the target classifier on source data weighted by how target-like the feature representation appears, given by $D(F(x))$. As in pseudo labeling, this weighting contrasts with weighting by the task classifier's softmax confidence. At test time, as in pseudo labeling, we can use the feature extractor followed by the target classifier for predictions. Note that this method is essentially selection adaptation \cite{daume2012course} but trained jointly and performed at a feature-level representation rather than the sample level.

\section{Experiments}
\begin{table*}
  \caption{Classification accuracy (source$\rightarrow$target) of the methods on benchmark computer vision datasets: MNIST, USPS, SVHN, MNIST-M, SynNumbers, SynSigns, and GTSRB. The strongest task vs. domain confidence on each dataset for each method is highlighted in bold (if not the same). The best-performing method in each column is underlined. The last method (italicized) is the one we propose in this paper. This method performs best on average.}
  \label{tab:results}
  {\renewcommand{\arraystretch}{1.3}
  \begin{tabular}{cccccccc}
    \toprule
    Method & MN$\rightarrow$US & US$\rightarrow$MN & SV$\rightarrow$MN & MN$\rightarrow$MN-M & $\text{Syn}_\text{N}\rightarrow$SV & $\text{Syn}_\text{S}\rightarrow$GTSRB & Average \\
    \midrule
    No Adaptation & 0.888 & 0.869 & 0.797 & 0.246 & 0.827 & 0.954 & 0.764 \\
    \hline
    DANN & 0.961 & 0.965 & 0.855 & 0.949 & 0.881 & 0.932 & 0.924 \\
    \hline
    Instance (task) & \textbf{0.960} & \textbf{0.964} & 0.831 & \textbf{0.943} & 0.876 & \textbf{0.936} & \textbf{0.918} \\
    Instance (domain) & 0.937 & 0.958 & \textbf{0.864} & 0.939 & \textbf{0.880} & 0.915 & 0.916 \\
    \hline
    Pseudo-NoAdv (task) & 0.968 & \textbf{0.965} & 0.741 & \textbf{0.789} & \underline{\textbf{0.909}} & \underline{\textbf{0.972}} & 0.891 \\
    Pseudo-NoAdv (domain) & \underline{\textbf{0.970}} & 0.960 & \textbf{0.869} & 0.721 & 0.901 & 0.963 & \textbf{0.897} \\
    \hline
    Pseudo-TaskC (task) & \textbf{0.962} & 0.970 & 0.861 & \underline{\textbf{0.986}} & \textbf{0.891} & \textbf{0.919} & 0.931 \\
    Pseudo-TaskC (domain) & 0.954 & \textbf{0.978} & \textbf{0.898} & 0.983 & 0.881 & 0.905 & \textbf{0.933} \\
    \hline
    Pseudo (task) & \textbf{0.952} & 0.972 & 0.873 & 0.985 & \textbf{0.898} & \textbf{0.934} & 0.936 \\
    \textit{Pseudo (domain)} & 0.950 & \underline{\textbf{0.979}} & \underline{\textbf{0.910}} & 0.985 & 0.894 & 0.924 & \underline{\textbf{0.940}} \\
    \bottomrule
  \end{tabular}
  } 
\end{table*}

We evaluate the method variations of no adaptation, DANN, instance weighting evaluated on the task or target classifiers (Instance-TaskC and Instance), pseudo labeling without the adversarial step that produces a domain-invariant feature representation (Pseudo-NoAdv), and pseudo labeling evaluated on the task and target classifiers (Pseudo-TaskC and Pseudo). Each instance weighting and pseudo labeling method is trained and evaluated both for weighting by the task classifier's softmax confidence (task) and by the discriminators confidence (domain).

We train these methods on popular computer vision datasets: MNIST \cite{lecun1998mnist}, USPS \cite{le1990handwritten}, SVHN \cite{netzer2011reading}, MNIST-M \cite{ganin2016jmlr}, SynNumbers \cite{ganin2016jmlr}, SynSigns \cite{moiseev2013evaluation}, and GTSRB \cite{Stallkamp-IJCNN-2011}. For MNIST$\leftrightarrow$USPS, we upscale USPS to 28x28 pixels to match MNIST using bilinear interpolation. For MNIST$\rightarrow$MNIST-M, we pad MNIST with zeros (before normalization) to be 32x32 pixels and convert to RGB to match MNIST-M. For SVHN$\rightarrow$MNIST, we pad MNIST to 32x32 and convert to RGB to match SVHN. The other datasets already have matching image sizes and depths.

For all experiments, we use the small CNN model used by Shu et al. \cite{shu2018vada} and for pseudo labeling use the task classifier architecture for the target classifier. We train each model 80,000 steps with Adam using a learning rate of 0.001 \cite{shu2018vada}, a batch size of 128 \cite{ganin2016jmlr}, the adversarial learning rate schedule from DANN \cite{ganin2016jmlr}, and a learning rate of 0.0005 for the target classifier. Target and source domain data is fed through the model in separate batches allowing for domain-specific batch statistics \cite{li2018,french2018iclr}. For model selection, we use 1000 labeled target samples from the training datasets as a holdout validation set. The reported accuracies are for the evaluation of each model (the target classifier for pseudo labeling and instance weighting methods, otherwise task classifier) on the testing sets that performed best on the holdout validation set. Thus, since in truly ``unsupervised'' domain adaptation situations we would not have any labeled target data, these results can be interpreted as an upper bound for how well these methods can perform \cite{wang2018domain}. Using some labeled target examples in this way is a common approach for tuning domain adaptation methods \cite{bousmalis2016nips,russo2018cvpr,wang2018domain,wei2018generative,carlucci2017autodial,kumar2018nips,shu2018vada}.

The results are summarized in Table~\ref{tab:results}. As indicated in these results, the proposed method of pseudo labeling with domain confidence performs the best. We can see that in all cases at least one of the adaptation methods improves over no adaptation, at least one of the pseudo labeling methods improves over instance weighting, and at least one of the pseudo labeling methods improves over DANN. On half of the datasets and on average, using adversarial training with pseudo labeling improves results. Typically, evaluating pseudo labeling on the target classifier is more effective than on the task classifier; though interestingly Pseudo-TaskC almost always outperforms DANN despite the task classifier never being updated by the pseudo labeling process. This indicates that pseudo labeling can improve the feature representation for the target domain. Finally, our primary goal was to determine if using a discriminator's confidence is more effective than a task classifier's softmax confidence, which is true on average for each of the pseudo labeling methods though not for instance weighting. Thus, these experiments appear to provide evidence that pseudo labeling with a domain discriminator's confidence may yield an improvement over a task classifier's softmax confidence.

\section{Conclusion}
In this paper, we investigated how to weight samples for pseudo labeling. We proposed using a discriminator's confidence rather than a task classifier's softmax confidence. The results of testing these methods on computer vision datasets provides insight into the possible benefit of using a discriminator not only for producing a domain-invariant feature representation but also for weighting samples for pseudo labeling.

Future work includes hyperparameter tuning either on the holdout set or with a method that does not require any labeled target data such as reverse validation \cite{ganin2016jmlr}. This method should be tested on additional datasets such as Office-31 \cite{saenko2010adapting} and on a greater variety of domain adaptation tasks. Additionally, we can  determine whether confidence thresholding \cite{french2018iclr} improves over confidence weighting. Finally, we can investigate theory behind selecting or weighting samples for pseudo labeling or instance weighting that may indicate why or when this method will work in addition to possible tweaks to yield improvements, such as possibly using a function of the discriminator's output rather than the probability directly, as is done in selection adaptation \cite{daume2012course}.

\bibliographystyle{ACM-Reference-Format}
\bibliography{bibliography}

\end{document}